# Finding the different patterns in buildings data using bag of words representation with clustering

Usman Habib, Gerhard Zucker
Energy Department, Sustainable Buildings and Cities
AIT Austrian Institute of Technology
Vienna, Austria
{usman.habib, gerhard.zucker}@ait.ac.at

*Abstract*—The understanding of the buildings operation has become a challenging task due to the large amount of data recorded in energy efficient buildings. Still, today the experts use visual tools for analyzing the data. In order to make the task realistic, a method has been proposed in this paper to automatically detect the different patterns in buildings. The K-Means clustering is used to automatically identify the ON (operational) cycles of the chiller. In the next step the ON cycles are transformed to symbolic representation by using Symbolic Aggregate Approximation (SAX) method. Then the SAX symbols are converted to bag of words representation for hierarchical clustering. Moreover, the proposed technique is applied to real life data of adsorption chiller. Additionally, the results from the proposed method and dynamic time warping (DTW) approach are also discussed and compared.

## I. Introduction

A lot of raw data is recorded during the monitoring of the energy efficient buildings [1]. In order to find the different aspect of the buildings performance the data is analyzed at later stages. The experts in the field usually analyze the data using different visualization tools [2]. The huge amount of data recorded makes it difficult for the experts to have a detailed performance analysis of buildings, thus making it hard to capture the different patterns, hence may lead to faults in the different components of building, reducing the energy efficiency.

The use of different data mining techniques can help in finding the different patterns in the buildings data, particularly clustering [3]–[5]. The automatic extraction of different patterns in large data set reduces the burden on experts in finding the different patterns in the data manually and helps in detailed analysis of the data. Therefore, the process of finding different patterns in the data can be feasible and less labor extensive [3], [4], [6].

In this paper an approach for automatically finding the different patterns in the building components operation has been proposed. In order to validate the outcomes, the proposed method has been applied to a data of adsorption chiller and compared to another approach called dynamic time warping (DTW). In first the ON/OFF cycle of the chiller is detected using the K-Means clustering algorithm, as the behavior of the chiller varies in these two different states. The patterns during the ON (operational) cycle is of greater importance for finding the performance of chillers and faults detection and diagnosis (FDD), therefore the data having ON cycle is considered in the proposed approach. Moreover, neglecting the OFF cycle will reduce the amount of data as well. The normalized ON cycles are discretized by using symbolic aggregate approximation (SAX). These discretized values are symbols or words. After transformation of the ON cycles to words, a histogram for each ON cycle is created called as bag of words representation (BoWR). Then the BoWR of ON cycles are clustered using hierarchical clustering. Furthermore, the results of the BoWR method with hierarchical clustering and DTW with hierarchical clustering are compared using cophenetic correlation. The cophenetic correlation demonstrates that the cluster tree have a strong correlation with the distances between objects in the distance vector [7].

The paper is structured as follows. In Section II the relevant research work is discussed. Section III describes the system design while Section IV explains the methodology of the proposed method. Section V explains the results and outcomes applied to real life data. Finally, the conclusion and future directions is given in Section VI.

## II. State of the Art

The advancement in sensors technology has made it feasible to record huge amount of data in commercial buildings. The huge amount of data storage makes it manually impossible to analyze it in detail. There are different tools used by experts in the field to visualize the data, which will further require manual analysis for performance of buildings. This process can be time extensive and there is always a chance to overlook some areas of interest [2].

There are different procedures available for faults detection and diagnosis (FDD) in buildings components e.g. HVAC (Heating, Ventilation and Air-Conditioning). Although, use of the earlier knowledge about the system can help in finding some of the faults with the first hand principles but still need more sophisticated techniques to find the different aspects of buildings performance. Black box models are available that are heavily dependent on the behavior or process of the system, which is generally captured from the historical data [8], [9]. Furthermore, the different machine learning algorithms can be used to detect faults in buildings by using the installed electricity consumption meter [10], [11]. Additionally, there are some parameters that can be useful for the prediction of electricity consumption for each of the HVAC component.

These parameters can be calculated by using multivariate analysis [12].

The key feature of data driven techniques in buildings is the focus on extracting knowledge from recorded data without detailed involvement of an expert. There are several popular unsupervised learning for extracting information. Such as, clustering can be used for finding the similar daily performance [3], [13], detecting the abnormal performance [14], and enhancing the performance optimization algorithms [15]. Moreover, wavelet transformations and clustering can be used in large scale for the classification of electrical demand profiles of buildings [16].

The data recorded for buildings are usually saved as time series data. There are different methods available for representing the time series data, thus will help in finding the similarity between the data having same behavior. The similarities can be found by using simple Euclidean distance between two time series data, but the problem will be even a slighter shift of data can lead to erroneous results [6]. There is another method called dynamic time warping (DTW) which overcomes the problem by using dynamic programming technique with defining the best alignment in data [17]. In [6] the comparison of algorithms (Euclidean, DTW, wavelets) with hierarchical cluster for finding different time series is done. They have also suggested the bag of patterns representation (BoPR) approach used for finding similarities in the time series data.

The bag of words model is used in several fields for classification [18]. The bag of words model can also be used with clustering to enhance the algorithm for finding the similarities between the time series data [6].

The time series data can be represented with a symbolic representation by using Symbolic Aggregate Approximation (SAX). The use of SAX improve the speed and usability of several analysis techniques. SAX is a category of Piecewise Aggregate Approximation (PAA) representation and is used extensively in various applications [19].

There are several clustering algorithms available, which can be used for finding the different states of buildings energy. For example, clustering can be used to detect the state of machine for ON/OFF cycle as data vary in these two different states. The [20] used the X-Means clustering algorithm for automatically detecting the system states (ON/OFF), to examine the operational data of adsorption. The ON/OFF state information can also be used for finding outliers in the data [21]–[23].

## III. SYSTEM DESIGN

This section describes the system that has been used for the proposed method in this paper. The brief description of adsorption chiller is given along with data description used for the analysis of the paper.

### A. Adsorption Chillers

For proof of concept anticipated in this paper, the data is taken from a solar cooling system with an adsorption chiller as defined in Figure 1. The naming conventions used in Figure 1 are according to the Task 38 of IEA solar heating and cooling program [24].

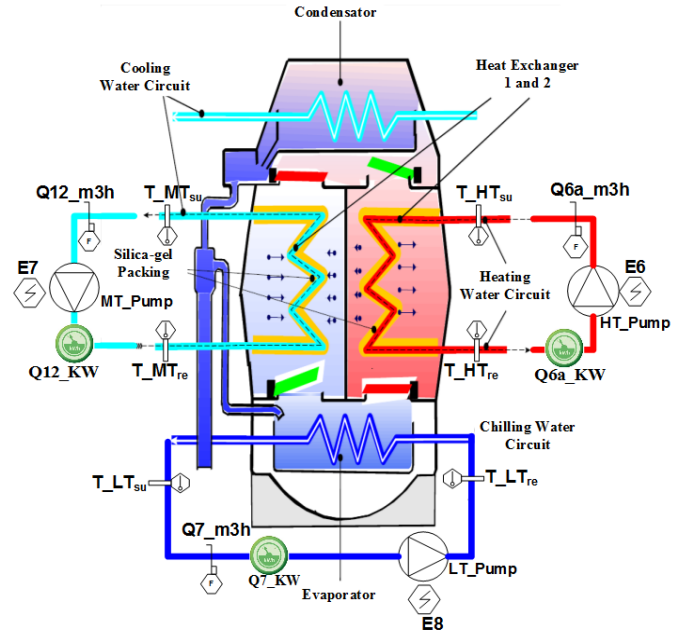

Figure 1: Adsorption chiller

The adsorption chiller operation can be described in the following steps [25]

1. The process of water evaporation is done in the lower chamber called evaporator, makes the water cool in Low Temperature (LT) cycle.

2. The evaporated water is adsorbed in the middle chamber, using silica-gel.

3. The adsorbed water is de-adsorbed with a certain heat provided from the Hot water (HT) cycle.

4. The de-adsorbed water is condensed and taken back to the evaporator.

### B. Data Description

The parameters under consideration are described in Table I given below. The naming convention of IEA Task 38 for solar and cooling is followed [24].

TABLE I : PARAMETERS DESCRIPTION

| Sensors | Description |
|---|---|
| E6 | High Temperature (HT) electricity consumption meter. |
| E7 | Medium Temperature (MT) electricity consumption meter. |
| E8 | Low Temperature (LT) electricity consumption meter. |
| Q6a_m3h | HT cycle Flow (water) reading |
| Q12_m3h | MT cycle Flow (water) reading |
| Q7_m3h | LT cycle Flow (water) reading |
| $T\_HT_{re}$ | HT cycle temperature on return side. |
| $T\_HT_{su}$ | HT cycle temperature on supply side. |
| $T\_MT_{re}$ | MT cycle temperature on return side. |
| $T\_MT_{su}$ | MT cycle temperature on supply side. |
| $T\_LT_{re}$ | LT cycle temperature on return side. |
| $T\_LT_{su}$ | LT cycle temperature on supply side. |
| Q6a_KW | HT cycle Energy consumption reading |
| Q12_KW | MT cycle Energy consumption reading |
| Q7_KW | LT cycle Energy consumption reading |

The data for the year 2010 is used for the proof of concept suggested in this paper, as can be seen in Figure 2. It is clear from Figure 2 that the cooling operation is recorded mostly during summer season. It is significant to notice that even during the summer season; there are days where little or no data is recorded due to communication line failures or some other unknown faults.

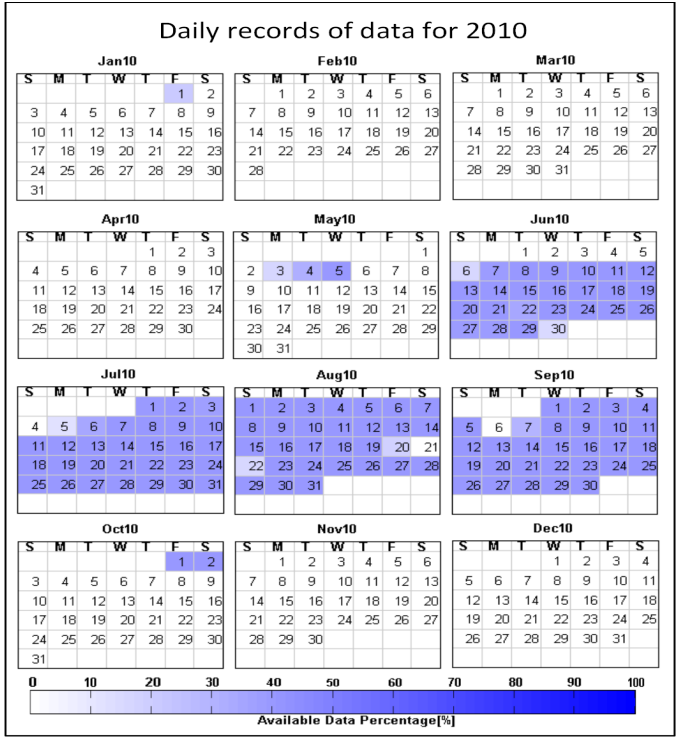

Figure 2: Data availability

## IV. METHODOLOGY

This section discuss the methodology proposed in this paper. The first step followed in the analysis of data is always the preprocessing and finding outliers. The data used has already been processed, therefore it can be used without the preprocessing step.

### A. ON State (Operational) detection using K-Means clustering

Generally the solar chillers are mostly in functional state during the summers. As the data varies a lot in the two states (ON/OFF) of chiller, therefore we can cluster the data in two clusters. The K-Means clustering with two cluster and Euclidean distance setting is used to detect the ON and OFF state. After the detection of ON/OFF state at each point of time, the consecutive ON state are marked as one ON cycle. The same process is done for all consecutive ON/OFF states. The ON status is of more importance than OFF cycle, for finding the performance of the chiller.

### B. Symbolic Aggregate Approximation (SAX) tranformation

After the detection of ON cycles, the ON cycle's data will be in the form as defined by the Eq.1.

$$C_i = \{S_1, S_2, S_3, \ldots \ldots \ldots \ldots S_N\} \quad \text{(Eq. 1)}$$

Where $C_i$ is the i[th] ON cycle and $S_k$ is the sensor value at time tick k. The data is normalized using Z-Score normalization before transformation to SAX symbols as given by the following Eq.2.

$$Z(C_i) = \frac{(S_k - \mu)}{\sigma} \quad \text{(Eq. 2)}$$

Here $S_k$ is the sensor data, $\mu$ represents the mean and $\sigma$ is the standard deviation of the data.

The normalized ON cycle data, $Z(C_i)$, is first broken down into *m* non-overlapping subsequences. This process is called as chunking, and the period length *P* is based on the application where it is used [3]. For this paper, *P* is chosen as 4 minutes due to the recording time interval of 4 minutes. The x-axis in Figure 3 represents the *P* as 4 minutes interval. Each piece is then further distributed into *Q* equal sized segments, e.g. alphabets a, b, c and d in Figure 3 are represented with different color regions. The symbol to each data point is assigned according the breakpoints. The number of break points taken for this research is 20, as with greater number of break point, the approximation will be quite near to the real data.

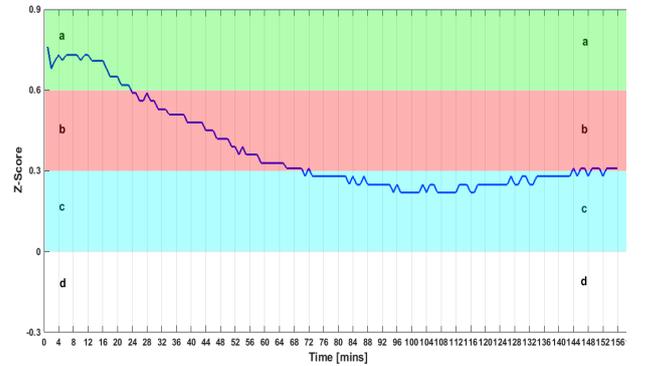

Figure 3: SAX Transformation of ON Cycle

### C. Bag of Words Representation (BoWR)

The SAX transformation will transform the time series data in the symbols. The bag of words representation can be explained by following Eq.3.

$$C_i = \{V_1, V_2, V_3, \ldots \ldots \ldots \ldots V_K\} \quad \text{(Eq. 3)}$$

Here K is the number of splits and $C_i$ is the vocabulary of the i[th] cycle, Whereas *V* is the word patterns of vocabulary given by Eq.4.

$$V_j = \text{Count}(V_j) \quad \text{(Eq. 4)}$$

Here *Count*() Function returns the total number of a given word in the vocabulary in a cycle. In order to handle the ON cycles with different time length the Eq.4 has been modified as Eq.5 given below, where *N* is the number of time ticks in ON cycle.

$$V_j = \frac{\text{Count}(V_j)}{N} \quad \text{(Eq. 5)}$$

The Table 2 shows the example of how the bag of words representation will denote each ON cycle. The benefit of using BoWR method is that it will reduce the dimensions of data as well.

TABLE 2: SAX VOCABULARY

| Time Series On Cycles ($C_i$) data | Cycle No | SAX Vocabulary | | | | | | | | |
|---|---|---|---|---|---|---|---|---|---|---|
| | | A | B | C | D | E | F | G | H | I | ... |
| | $C_1$ | 0 | 5 | 4 | 3 | 2 | 1 | 0 | 0 | 1 | ... |
| | $C_2$ | 1 | 9 | 7 | 4 | 5 | 3 | 1 | 5 | 2 | ... |
| | .... | .... | .... | .... | .... | .... | .... | .... | .... | .... | |
| | $C_N$ | ... | .... | .... | .... | .... | .... | .... | .... | .... | |

### D. Key Steps in Methodology

The main steps of the methodology are as following

1. As the data is preprocessed, therefore the focus is on next steps. The first step in to find the ON (operational) cycles in data by using K-Means algorithm.
2. Transform the ON cycle's data to symbolic data using SAX transformation method.
3. Create BoWR of the symbols of each ON cycle.
4. Cluster the BoWR using hierarchical clustering for finding the different performance of chiller.

## V. RESULTS

The methodology is applied to real life data. In this section the different results are discussed as following

### A. ON Cycle Detection with K-Means Algorithm

The results of the K-Means clustering can be seen in Figure 4. The Figure 4 validates the ON/OFF status through K-Means clustering algorithm. The dashed line (red) in the figure shows the ON/OFF cycles of the chiller. It can be observed from the behaviour of the temperatures at the LT, MT and HT cycle are reacting according to the detected ON/OFF state. The dotted rectangle shows one ON cycle of the chiller between 21-07-2010 and 22-07-2010. It is clear from the Figure 4 that during the detected ON cycle, the LT temperatures decreases showing the cooling operation. Whereas, at the same time the temperatures increase in the HT and MT cycle of the chiller. This simultaneously change in temperature, is a clear signal that the chiller is in operational mode, which has also been detected by the proposed method of K-Means clustering.

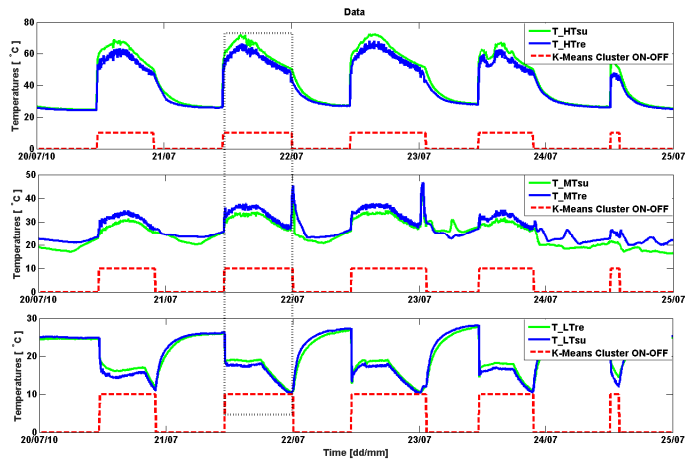

Figure 4: K-Means Clustering for Finding ON/OFF cycle

The Figure 5 shows the different ON cycles considered for next steps. The ON cycles are different in length as well as different in performance of chillers at these time stamps. The Carnot coefficient of performance ($COP_{carnot}$) [26] is used as given in Eq.6, whereas the thermal coefficient of performance ($CoP_{therm}$) can be calculated using Eq.7. The average $Efficiency$ for chiller is calculated for each cycle by using Eq.8.

$$COP_{carnot} = \left(\frac{T\_HT_{su} - T\_MT_{su}}{T\_HT_{su}}\right) * \left(\frac{T\_LT_{re}}{T\_MT_{su} - T\_LT_{re}}\right) \quad \text{(Eq. 6)}$$

$$CoP_{therm} = \frac{Q7\_KW}{Q6a\_KW} \quad \text{(Eq. 7)}$$

$$Efficiency = \frac{COP_{carnot}}{CoP_{therm}} \quad \text{(Eq. 8)}$$

In Figure 5, the cycle 1, 2 and 3 have the best chiller efficiency which is above 50%. Cycle 4 and 5 having efficiency between 30% and 50% whereas, cycle 6 and 7 have the low efficiency of lower than 30%. One of the reasons for low efficiency can be the shorter period of cooling load. In Figure 5 cycles 6 and 7 (dash lines) are representing the low chiller efficiency, while, cycles 4 and 5 (dot lines) represent the cycles with average efficiency and cycles 1, 2 and 3 have good efficiency.

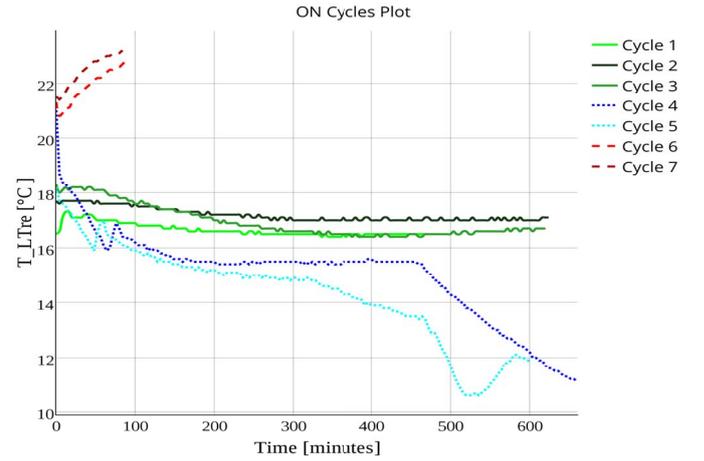

Figure 5: ON Cycle Plot

### B. Symbolic Aggregate Transformation (SAX) and Bag or Words Representation

For SAX model the chunk size is taken as 4 minutes as the data is recorded at an interval of 4 minutes. The breaking points for chunks is taken as 20. The bag of words representation (BoWR) can be seen in Figure 7. The BoWR is represented by the histograms in the second column of the Figure 7. The green colour is used for good chiller efficiency cycles, whereas, blue denotes the average efficiency and red colour histograms is reflecting the bad efficiency cycles. It can be observed from the Figure 7 that green histograms are in mid and are similar to each other as compared to others. The same can be noticed for the average and bad efficiency cycles. This bag of word representation helps the clustering to identify the cycles.

## C. Hierarchical Clustering

The hierarchical clustering is applied with Dynamic time warping and proposed method of bag of words representation (BoWR) method. Figure 6 shows the results of clustering with DTW and Figure 7 is describing the results with BoWR. Both methods have classified the cycles correctly. The first advantage of BoWR over DTW is the dimension reduction, as each cycle is represented with significantly less data. Secondly, for comparing the clustering performance with both methods the cophenetic coefficients [7] has been calculated for hierarchical clustering. The cophenetic correlation demonstrates that the cluster tree have a strong correlation with the distances between objects in the distance vector. Table 3 shows the cophenetic coefficients with different hierarchical clustering methods for BoWR and DTW techniques. The BoWR has strong correlation with distance with all other objects in all clustering method. The best results for BoWR are attained with 'Average' method for hierarchical clustering.

TABLE 3: COPHENETIC COEFFICIENTS

| No. | Clustering Methods | BoWR | DTW |
|---|---|---|---|
| 1 | Average | 0.9897 | 0.0375 |
| 2 | Centroid | 0.9851 | 0.037 |
| 3 | Complete | 0.9753 | 0.035 |
| 4 | Median | 0.9803 | 0.0363 |
| 5 | Single | 0.9848 | 0.0414 |
| 6 | Ward | 0.9835 | 0.0363 |
| 7 | Weighted | 0.9888 | 0.0368 |

## VI. CONCLUSION

A bag of word representation (BoWR) method with hierarchical clustering is proposed in this paper for finding the different patterns in the chiller performance. First a K-Means clustering is used to find the ON (operational) cycles of the chiller. These ON cycles are represented with symbols by using symbolic aggregate approximation (SAX) method. Furthermore, the symbolic representation is transformed to BoWR, which is provided to hierarchical clustering for detection of bad, average and good Carnot coefficient of performance (COP).

The Dynamic time warping (DTW) is also implemented with hierarchical clustering and results are compared with the proposed method. The important points during comparison are following:

- The cophenetic coefficients have shown that the BoWR has produced better results as compared to DTW.

- The other benefit of using BoWR is the reduction in dimension as a large amount of ON cycle is represented with less details.

In future, the current research can be used in the field of automatic detection and diagnostics of faults (FDD) in buildings, as the current research helps in finding the different performance patterns. This would help the experts in the field to look only for those areas where the performance is bad. The further research is needed to find intelligent way of diagnosing the faults

ACKNOWLEDGMENT

This work was partly funded by the Austrian Funding Agency in the funding programme e!MISSION within the project "extrACT", project number 838688.

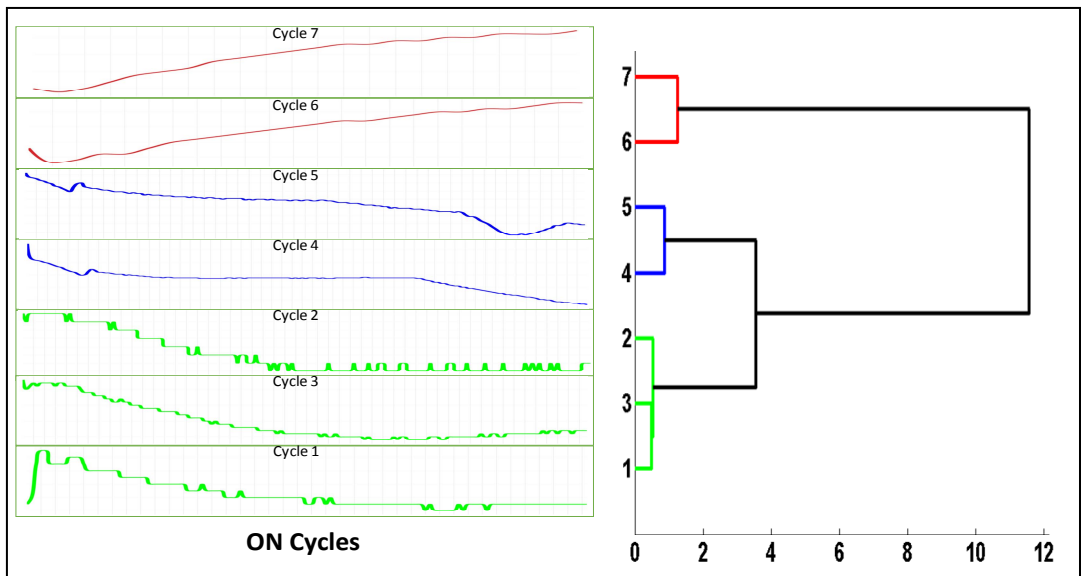

Figure 6: Clustering of ON cycle data with Dynamic Time Warping (DTW)

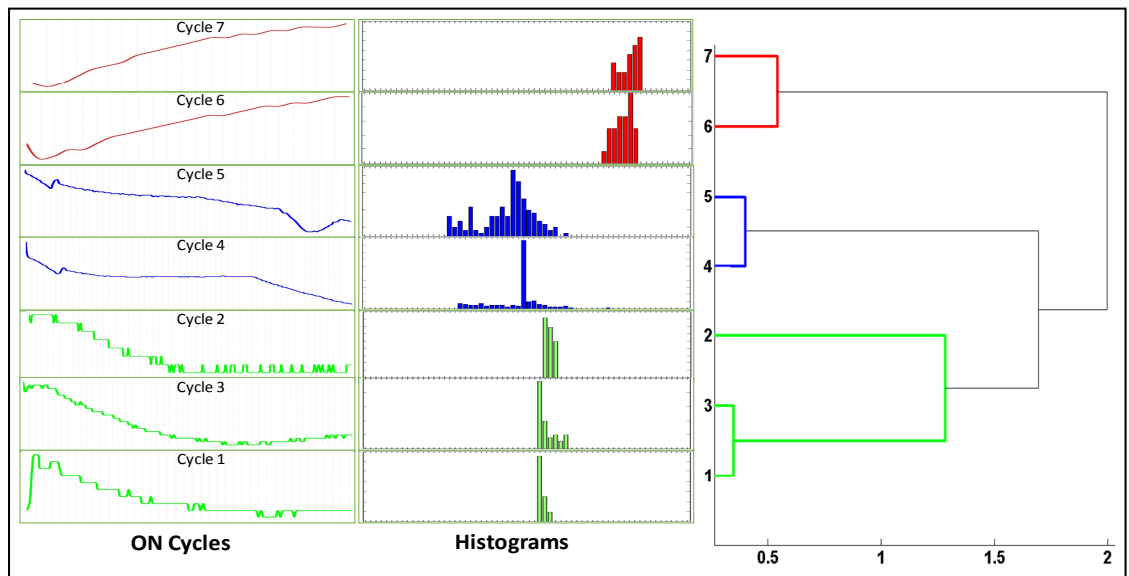
Figure 7: Clustering of ON cycles with bag of words Representation method